\documentclass[10pt,twocolumn]{article}

\usepackage[utf8]{inputenc}
\usepackage[T1]{fontenc}
\usepackage{amsmath,amssymb,amsthm}
\usepackage{booktabs}
\usepackage{array}
\usepackage{xcolor}
\usepackage[margin=0.75in,columnsep=0.25in]{geometry}
\usepackage[hidelinks]{hyperref}
\usepackage{microtype}
\usepackage{graphicx}
\usepackage{tikz}
\usetikzlibrary{arrows.meta,positioning,calc}

\newcommand{\code}[1]{\texttt{\detokenize{#1}}}
\newcommand{\fma}{\textsc{fma}}
\newcommand{\rne}{\mathrm{RNE}}
\newtheorem{lemma}{Lemma}
\theoremstyle{definition}

\theoremstyle{remark}
\newtheorem{remark}{Remark}

\title{\bf Formally Verified Synthesizable Floating-Point Data Types in ARCH HDL}

\author{%
  Shuqing Zhao\\
  \texttt{shuqing.zhao@gmail.com}%
}
\date{\today}

\begin{document}
\maketitle

\begin{abstract}
We report the design and end-to-end verification of first-class IEEE-754 binary32
(FP32) and bfloat16 (BF16) arithmetic for ARCH, a hardware description language
intended to be generated by language models from natural-language specifications.
Every floating-point operator --- comparisons, conversions, add, sub, multiply,
and fused multiply--add, in both formats --- is described once in a host-language
embedded DSL against a single bit-vector intermediate representation, which is
rendered three ways from one source: to synthesizable SystemVerilog (the
hardware), to SMT-LIB (a formal model), and to Lean~4 (a structured proof model).
Because the three back-ends consume one description and one linearization, the
hardware, the formal model, and the proof model cannot drift apart
structurally: they are the same dataflow graph printed in three syntaxes.
What remains between them is the per-node syntax table of roughly twenty
printers; on the hardware side even that is machine-checked --- Yosys
independently reads the emitted SystemVerilog and a solver proves it
equivalent to the SMT model, for all operators. We verify the
operators with a hybrid strategy matched to the
solver-tractability frontier: the \emph{multiplier-free} operators (all six FP32
comparisons and bounded add/sub over the full $2^{64}$ input space, the format
conversions, and all \emph{binary} BF16 arithmetic over $2^{32}$) are proved
equivalent to the SMT-LIB \texttt{FloatingPoint} theory exhaustively; the
\emph{multiplier-bearing} operators (FP32 \code{mul} and \fma{}), which are
SAT-hard for any bit-blaster, are machine-proved correctly rounded in Lean by
algebraic lifting, with zero \code{sorry}. (The one deliberate exception: the
BF16 \fma{} is an FP32-accumulating fusion --- the Tensor-Core/TPU convention
--- machine-characterized as exactly that, not claimed correctly rounded.)
Physical characterization (open-source
Yosys synthesis and OpenSTA timing) then exposed the \fma{} as the timing outlier:
its exact-wide $470$-bit aligned adder is a single combinational cone that our
synthesis flow cannot pipeline (ABC retiming moves no registers through it). We
therefore reimplemented the \fma{} as a bounded $98$-bit guard /
round / sticky datapath that pipelines cleanly to $268$\,MHz --- and proved, in
Lean and \emph{sorry}-free, that the bounded datapath is \emph{bit-identical} to
the exact-wide reference over all $2^{96}$ inputs, so it inherits the reference's
proven correct rounding without re-engaging the multiplier. The equivalence proof
is tractable precisely because the shared multiplier appears on both sides and
cancels. We give the architecture, the two-tier proof methodology, the synthesis
results, the trust base (the standard three axioms plus \code{bv_decide}'s
native-checked LRAT-certificate axioms), and an independent adversarial audit.
\end{abstract}

\section{Introduction}

ARCH\footnote{\url{https://github.com/arch-hdl-lang/arch-com}} is a
purpose-built hardware description language for micro-architecture work,
designed so that the language surface can be generated correctly by language
models from natural-language hardware descriptions, and compiled to deterministic,
readable SystemVerilog. A language meant to describe real datapaths needs
first-class floating-point: not a library bolted on, but FP32~\cite{ieee754} and
BF16~\cite{bf16} as built-in types with operators that are (i) \emph{correctly rounded} to IEEE-754, (ii)
\emph{synthesizable} to efficient RTL, and (iii) \emph{trustworthy}, since
machine-generated hardware most needs machine-checked guarantees. These three
goals are in tension. Correct rounding is a precise but subtle specification.
Exhaustive formal equivalence checking hits a wall at the one operator that
matters most for throughput --- the fused multiply--add --- because its
significand multiplier makes the miter SAT-hard~\cite{bieremiters,amulet} (a
\emph{miter} ties the two circuits under comparison to shared inputs and XORs
their outputs, reducing equivalence to unsatisfiability of that XOR). And the most obviously-correct
\fma{} datapath (compute the exact wide product-sum, then round) is the worst for
timing.

This paper reports how we resolved all three for ARCH, and the path is the
contribution as much as the endpoint:

\begin{enumerate}
\item \textbf{One source, three back-ends.} Each operator is defined once against
  a shared bit-vector IR --- a DAG whose subterms are shared by construction;
  three renderers emit synthesizable SystemVerilog,
  SMT-LIB, and Lean~4 from that one description and one linearization
  (Section~\ref{sec:arch}). The hardware, the SMT model, and the proof model
  are \emph{structurally} the same artifact in three syntaxes; the
  SystemVerilog printer's semantic correspondence is machine-checked by a
  Yosys-to-SMT miter, the Lean printer's by a byte-identical regeneration
  audit (Section~\ref{sec:limitations}).
\item \textbf{A two-tier proof methodology at the tractability frontier.}
  Multiplier-free operators are discharged \emph{exhaustively} by SMT against the
  IEEE-754 \texttt{FloatingPoint} theory; multiplier-bearing operators are
  \emph{structurally} proved in Lean by lifting the bit pattern to an algebraic
  value, so the $24\times24$ array never gets bit-blasted
  (Section~\ref{sec:verif}).
\item \textbf{Timing-driven design, verified after the fact.} Synthesis and static
  timing (Section~\ref{sec:synth}) show the exact-wide \fma{} is the outlier and
  resisted retiming (ABC moved no registers through the cone in any tested
  configuration); we redesign it as a bounded sticky-fold datapath
  (Section~\ref{sec:sticky}) that pipelines to $268$\,MHz, and prove it
  \emph{bit-identical} to the exact-wide reference over all $2^{96}$ inputs
  (Section~\ref{sec:equiv}), so the fast datapath inherits the reference's proven
  correct rounding. The equivalence is provable without re-engaging the
  multiplier, because the shared multiplier cancels.
\end{enumerate}

\noindent The result is a floating-point feature whose every operator is either
exhaustively or structurally verified against its stated contract (for the BF16
\fma{}, an FP32-accumulating fusion, that contract is stated rather than
correct rounding; Section~\ref{sec:bf16fma}), on a model that is provably the
emitted RTL, with a clean axiom base and an independent audit
(Section~\ref{sec:trust}).

\section{Background}\label{sec:background}

\paragraph{Correct rounding.} For a destination format with $p$-bit significand,
a correctly rounded operation returns $\rne(v)$, where $v$ is the \emph{exact} real
result and $\rne$ is round-to-nearest, ties-to-even~\cite{ieee754}. For \fma{},
$v = a\cdot b + c$ formed without intermediate rounding. ``Correctly rounded'' is
thus two claims: form the exact value, and round it.

\paragraph{Guard / round / sticky.} Hardware keeps the result significand plus
three summary bits: guard (the first bit below the kept significand), round (the
next), and sticky (the OR of all lower bits). RNE on a GRS-encoded value is exact
--- round up iff $\text{guard}\wedge(\text{round}\vee\text{sticky})$, ties to even
--- so the engineering problem is producing the GRS bits without materializing the
full exact value.

\paragraph{The multiplier wall.} After special values are dispatched, equivalence
of an arithmetic circuit to its specification is a quantifier-free bit-vector (and
floating-point) formula. Bit-blasting turns the $24\times24$ significand
multiplier into a quadratic gate network whose conjunctive normal form is
intractable across the binary32 input space. Multiplier miters are canonically
hard SAT instances --- a standard benchmark family for exactly this
reason~\cite{bieremiters}, with lower-bound analyses of why solvers struggle
on nonlinear integer arithmetic~\cite{beameliew} --- hard enough that
state-of-the-art multiplier verification abandons plain SAT for computer
algebra~\cite{amulet}. Consistently, z3, cvc5, and the Lean tactic
\code{bv_decide}~\cite{bvdecide} all stall on \code{mul} and \fma{} here. Everything
without a multiplier --- comparisons, conversions, and bounded add/sub whose
datapath stays narrow --- is exhaustively decidable; the multiplier-bearing
operators are not. Our methodology is organized around exactly this frontier.

\section{Single-source architecture}\label{sec:arch}

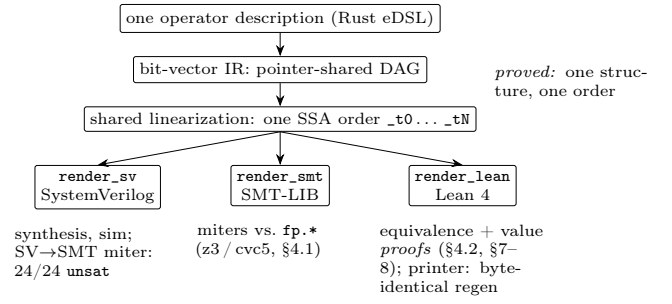
\begin{figure}[t]
\centering
\resizebox{\columnwidth}{!}{%
\begin{tikzpicture}[
  font=\scriptsize, >=Stealth,
  box/.style={draw, rounded corners=1pt, align=center, inner sep=2.5pt},
  ann/.style={font=\scriptsize, align=left, text width=2.5cm}]
  \node[box] (desc) {one operator description (Rust eDSL)};
  \node[box, below=3.5mm of desc] (ir) {bit-vector IR: pointer-shared DAG};
  \node[box, below=3.5mm of ir] (lin) {shared linearization: one SSA order \texttt{\_t0}\,\ldots\,\texttt{\_tN}};
  \node[box, below=5mm of lin, xshift=-2.75cm] (sv) {\texttt{render\_sv}\\ SystemVerilog};
  \node[box, below=5mm of lin] (smt) {\texttt{render\_smt}\\ SMT-LIB};
  \node[box, below=5mm of lin, xshift=2.75cm] (lean) {\texttt{render\_lean}\\ Lean~4};
  \node[ann, below=1.5mm of sv] {synthesis, sim;\\ SV$\to$SMT miter:\\ 24/24 \texttt{unsat}};
  \node[ann, below=1.5mm of smt] {miters vs.\ \texttt{fp.*}\\ (z3\,/\,cvc5, \S4.1)};
  \node[ann, below=1.5mm of lean] {equivalence + value \emph{proofs} (\S4.2, \S7--8); printer: byte-identical regen};
  \draw[->] (desc) -- (ir);
  \draw[->] (ir) -- (lin);
  \draw[->] (lin.south) -- (sv.north);
  \draw[->] (lin.south) -- (smt.north);
  \draw[->] (lin.south) -- (lean.north);
  \node[font=\scriptsize, anchor=west, text width=2.3cm, align=left]
    at ($(lin.east)+(0.15,0.55)$)
    {\emph{proved:} one structure, one order};
\end{tikzpicture}}
\caption{One description, three back-ends. Everything above the fork is a
single shared artifact (structural identity is by construction); the residual
trust is the per-node syntax table below the fork; the SystemVerilog side is
discharged mechanically by the Yosys-to-SMT miter (all $24$ operators
\texttt{unsat}, \S\ref{sec:limitations}), the Lean side by the
byte-identical regeneration audit.}
\label{fig:toolchain}
\end{figure}

Each operator is built once, in Rust, against a pointer-shared bit-vector IR
(Figure~\ref{fig:toolchain}): a
directed acyclic graph of nodes (\code{add}, \code{sub}, \code{mul}, \code{ite},
\code{extract}, \code{concat}, \code{zero_extend}, the bit-vector comparisons, and
named calls), where a subterm the builder reuses is one node, not a copy. One
shared \emph{linearization} pass flattens that DAG into a straight-line
program: a single topological sequence of temporary assignments \code{_t0},
\code{_t1}, \ldots\ in single-static-assignment form --- each temp defined
once, referring only to earlier temps, each shared node emitted exactly once.
The three renderers walk that one numbered sequence (Table~\ref{tab:renderers})
and differ only in how a line is spelled, via the per-node syntax table (a
\code{Bin::Add} node prints \code{+} in SystemVerilog, \code{bvadd} in
SMT-LIB, and \code{+} on \code{BitVec} in Lean); consequently \code{_t44}
names the \emph{same node} in the hardware, the SMT model, and the proof
model.

\begin{table}[t]
\centering
\small
\setlength{\tabcolsep}{4pt}
\begin{tabular}{lll}
\toprule
renderer & output & consumer\\
\midrule
\code{render_sv}   & SystemVerilog & \code{arch build}\\
\code{render_smt}  & SMT-LIB2 \code{define-fun}s  & z3 / cvc5\\
\code{render_lean} & Lean~4 \code{BitVec} \code{def}s & this proof\\
\bottomrule
\end{tabular}
\caption{One bit-vector IR, three renderers. Sharing the IR \emph{and} the
linearization, the renderers cannot disagree on structure --- only on the
per-operator syntax table.}
\label{tab:renderers}
\end{table}

Because all three consume the same IR and the same linearization, they cannot
disagree on sharing, control flow, or node order; the Lean proof model is a
generated snapshot of \code{render_lean}, and we confirm
(Section~\ref{sec:trust}) it is byte-identical to a fresh regeneration --- so the
proof is about the emitted artifact, not a paraphrase. The only way the three
artifacts \emph{can} diverge is a wrong entry in the syntax table --- a
mistranslation of a single node kind, which would appear uniformly in every
operator using that node, not as an operator-specific bug. That failure mode
is what Section~\ref{sec:limitations} scopes as the residual renderer trust
--- and what the Yosys-to-SMT miter now discharges mechanically for the
SystemVerilog renderer. The operator set spans the
six FP32 comparisons; FP32 \code{add}, \code{sub}, \code{mul}, \fma{}; the
conversions FP32$\to$BF16 (narrowing, RNE), BF16$\to$FP32 (widening, exact), and
FP32$\to$signed/unsigned integer (toward-zero, saturating); and the BF16
comparisons and arithmetic, which route through the FP32 datapath. NaN
canonicalization follows a selectable profile (e.g.\ the RISC-V vs.\ CUDA
conventions).

\section{Verification methodology}\label{sec:verif}

\subsection{Tier 1 --- exhaustive SMT for the multiplier-free operators}

The SMT renderer emits each operator as a \code{define-fun}; a miter asserts the
negation of equivalence to the SMT-LIB \texttt{FloatingPoint}
theory~\cite{smtfp} --- the standardized formalization of IEEE-754 RNE, the same
value semantics Berkeley SoftFloat implements. An \code{unsat} verdict means the
emitted RTL operator matches the theory's RNE \emph{value} semantics --- the
result bit pattern, with our canonical-NaN choice pinned explicitly in the
miter --- over its entire input space. We discharge, exhaustively:
\begin{itemize}
\item all six FP32 comparisons and the conversions over their full domains (the
  $2^{64}$ compare space; FP32$\to$BF16 narrowing over $2^{32}$; BF16$\to$FP32
  widening over $2^{16}$; FP32$\to$integer for in-range inputs, the partial-function
  boundary documented rather than silently claimed);
\item FP32 \code{add} and \code{sub} versus \code{fp.add}/\code{fp.sub} over all
  $2^{64}$ inputs. The decisive factor is datapath width, not input space: the
  bounded adder keeps the aligned magnitude near $56$ bits (no multiplier), so the
  bit-blasted miter stays small;
\item all \emph{binary} BF16 arithmetic (\code{mul}, \code{add}, \code{sub}, and
  the six compares) over the full $2^{32}$ input space. BF16's small input space
  makes even the multiplier-bearing binary operators tractable, even though they
  route through the FP32 datapath; cross-checked across solvers. The ternary
  BF16 \fma{} (a $2^{48}$ space) is deliberately \emph{not} on this list:
  it is FP32-accumulating, and its distinct contract is stated in
  Section~\ref{sec:bf16fma}.
\end{itemize}

\paragraph{The exact contract.} ``IEEE-754'' in this paper means
\emph{result-bit value semantics under a fixed rounding regime}, and no more.
Rounding is statically RNE for arithmetic and format conversions (toward-zero
for the saturating float-to-integer path); there is no dynamic rounding-mode
input, and no status flags --- \code{invalid}, \code{inexact},
\code{overflow}, \code{underflow} are not computed. NaN handling is by
explicit canonical profile: IEEE-754 and SMT-LIB both abstract NaN payloads, so
each miter's NaN branch pins our canonicalized output rather than appealing to
the theory. Comparisons implement the quiet predicates (\code{fp.eq} and
kin); there are no signaling variants and hence no invalid-operation side
channel. Float-to-integer is \emph{saturating} with defined NaN and
out-of-range results at the ARCH level; SMT-LIB leaves
\code{fp.to_sbv}/\code{fp.to_ubv} unspecified there, so the miter covers
exactly the in-range domain where the theory is defined, and the
saturation/NaN behavior is a documented ARCH contract validated by directed
simulation rather than by the proof. Finally, all three models are
\emph{two-state}: the IR and its SMT and Lean renderings have no
\code{X}/\code{Z}, so SystemVerilog four-state behavior (\code{X}
propagation, high impedance) is outside the verified contract, which speaks
to fully driven binary inputs.

\subsection{Tier 2 --- structured Lean for the multiplier wall}

FP32 \code{mul} and \fma{} carry the $24\times24$ multiplier and time out for any
bit-blaster. A structured prover clears the wall by \emph{not} bit-blasting.
Following the Flocq / FloVer methodology~\cite{flocq,flover}, the proof lifts the
bit pattern to an algebraic $(\text{sign},\text{significand},\text{exponent})$ view
over a rational value and shows the operator computes $\rne(a\cdot b\,[+\,c])$
structurally: the $24\times24$ array collapses to a single \code{Nat} multiply,
whose algebra is elementary. The development factors into a reusable
\emph{rounder} characterization --- that the shared round-to-nearest-even encoder
finds the true leading bit (a count-leading-zeros / most-significant-bit bridge),
performs an exact round-trip, and satisfies $1\cdot x = x$ --- and per-operator
reductions on top of it. For \fma{} this yields \code{arch_round470_correct} and a
\emph{sorry}-free proof that the finite FP32 \fma{} is correctly rounded against a
value-level RNE specification. The development is dependency-free (Lean core
\code{BitVec} and \code{bv_decide}; no Mathlib).

\subsection{BF16 \fma{}: an honest exception}\label{sec:bf16fma}

The BF16 \fma{} is intentionally a \emph{fused f32-accumulate}: one correctly
rounded FP32 \fma{} followed by a narrowing rounding FP32$\to$BF16. It is therefore
\emph{not} a correctly rounded BF16 \fma{}: the second rounding is not innocuous,
and it differs from $\rne_{\text{bf16}}(a\cdot b + c)$ on a small fraction of finite
inputs, always by one unit in the last place. This is the FP32
accumulation-format convention of NVIDIA tensor cores~\cite{ampere}, Google
TPUs~\cite{tpubf16}, and BF16 training practice~\cite{bf16} --- a convention
about the accumulation format only; vendor implementations are not otherwise
bit-identical (TPU subnormal handling differs, for instance). The operator is
separately machine-characterized as ``the narrowing of the correctly rounded
FP32 \fma{}'' rather than claimed as correctly rounded. An earlier ``innocuous double
rounding'' argument for it was a round-to-nearest fallacy; we record the corrected
characterization.

\subsection{Differential backstop} The same IR also drives a differential
campaign against Berkeley SoftFloat~\cite{softfloat}, the de-facto reference
IEEE-754 implementation, on directed corner vectors (signed zeros and infinities,
quiet and signaling NaNs, subnormal and normal extrema, exact ties, overflow and
underflow boundaries) and large randomized vectors, providing an empirical
cross-check independent of the SMT theory.

\section{Physical characterization}\label{sec:synth}

We synthesized the emitted SystemVerilog of every arithmetic and format-conversion
operator through an open-source flow (Yosys for logic synthesis, OpenSTA for
static timing), reporting maximum frequency as
$f_{\max}=1000/\text{(critical-path delay in ns)}$ against a virtual clock.
Table~\ref{tab:peropfmax} gives the per-operator combinational result on
Nangate45 standard cells (typical corner); the six comparisons and the
saturating integer conversions carry no rounding chain and are not
characterized here. The complete flow --- the Yosys scripts, the buffered
\code{abc} repair script of Remark~\ref{rem:buffering}, the OpenSTA scripts,
tool versions (Yosys 0.67, OpenSTA 3.1.0), and the Liberty file's provenance
by SHA-256 --- is archived in the repository
(\code{tests/fp_v1/synth/nangate45/}), with a reproduction check: the archived
seven-stage \fma{} source re-measures at $268.0$\,MHz, and a representative
combinational operator regenerated from the current compiler lands within
$3\%$ of its published figure (compiler-side RTL drift, not flow drift).

\begin{remark}[Fanout buffering matters]\label{rem:buffering}
Yosys's default \code{abc -liberty} mapping performs \emph{no} fanout buffering
or post-mapping gate sizing, so high-fanout nets accumulate hundreds of
femtofarads and the reported $f_{\max}$ is dominated by a handful of overloaded
drivers rather than by logic depth. All numbers in this paper therefore use an
explicit repair pass in the \code{abc} script (\code{buffer -N 8; upsize;
dnsize}), which eliminates every max-capacitance violation. The effect is not a
refinement but a correction: the exact-wide \fma{} measures $45$\,MHz unbuffered
and $160$\,MHz buffered --- a $3.5\times$ artifact. (AIG optimization is
\code{dch -f}, with \code{dc2} substituted for the two BF16 add/sub netlists on
which \code{abc}'s \code{dch} aborts; the mapping and repair stages, which carry
the timing result, are identical for all operators.)
\end{remark}

\begin{table*}[t]
\centering
\begin{tabular}{lrr@{\qquad}lrr}
\toprule
operator & $f_{\max}$ & cells & operator & $f_{\max}$ & cells\\
\midrule
BF16$\to$FP32 (widen)  & 7456 & 25   & FP32 \code{sub} & 318 & 2288 \\
FP32$\to$BF16 (narrow) & 3997 & 94   & FP32 \code{mul} & 276 & 6124 \\
BF16 \code{mul}        & 468  & 1664 & BF16 \fma{}     & 185 & 4872 \\
BF16 \code{sub}        & 420  & 1315 & \textbf{FP32 \fma{} (exact-wide)} & \textbf{160} & \textbf{19152} \\
BF16 \code{add}        & 406  & 1323 & FP32 \fma{} (sticky fold, comb) & 180 & 10805 \\
FP32 \code{add}        & 320  & 2263 & \\
\bottomrule
\end{tabular}
\caption{Per-operator combinational $f_{\max}$ (MHz) and mapped cell count,
Nangate45 standard cells (typical corner), Yosys $+$ OpenSTA, buffered flow
(Remark~\ref{rem:buffering}). The widen/narrow conversions are near-wire; every
arithmetic operator clears $276$\,MHz --- except the \fma{}s, and the exact-wide
FP32 \fma{} is the clear outlier: slowest at $160$\,MHz and, at $19$k cells,
$3\times$ the area of the multiplier.}
\label{tab:peropfmax}
\end{table*}

The exact-wide \fma{} is not merely the slowest: it does not \emph{pipeline}. Its
$470$-bit aligned adder is a single combinational cone, and ABC's register
retiming moves no flops into such a cone at all (we verified this directly: under
every retiming configuration we tried, the register count and positions are
unchanged), while a hand-inserted three-stage split of the $470$-bit datapath
yielded no improvement either. This is the design problem the next section
addresses. (The exact-wide \fma{} is, however, the simplest \emph{specification}:
it computes the exact value and rounds once, and is the object the equivalence
proof targets.)

\section{The bounded sticky-fold \fma{}}\label{sec:sticky}

\begin{figure*}[t]
\centering
\begin{tikzpicture}[font=\scriptsize, >=Stealth, yscale=0.9]
  \node[anchor=west, font=\scriptsize\bfseries] at (0,2.65) {exact-wide reference (470\,b): alignment exact, nothing dropped};
  \draw (0,1.75) rectangle (13.2,2.25);
  \fill[blue!14] (1.1,1.75) rectangle (4.4,2.25);
  \draw (1.1,1.75) rectangle (4.4,2.25);
  \node at (2.75,2.0) {hi $\cdot\,2^{\Delta}$ \ (48\,b)};
  \fill[orange!22] (9.9,1.75) rectangle (13.2,2.25);
  \draw (9.9,1.75) rectangle (13.2,2.25);
  \node at (11.55,2.0) {lo \ (48\,b)};
  \draw[<->] (4.4,1.55) -- (9.9,1.55) node[midway, below=-1pt] {gap $\Delta \le 421$ --- every lo bit retained};
  \node[anchor=west, font=\tiny] at (0.05,2.0) {$2^{469}$};
  \node[anchor=east, font=\tiny] at (13.15,1.45) {$2^{0}$};
  \node[anchor=west, font=\tiny, text width=2.2cm] at (13.35,2.0) {add / $|{-}|$,\\ round once};
  \node[anchor=west, font=\scriptsize\bfseries] at (0,0.9) {sticky-fold (98\,b): same hi, lo survives only down to the guard; the rest ORs into one sticky bit};
  \draw (0,0.0) rectangle (13.2,0.5);
  \fill[blue!14] (0,0.0) rectangle (6.5,0.5);
  \draw (0,0.0) rectangle (6.5,0.5);
  \node at (3.25,0.25) {hi at weight $2^{49}$ \ (48\,b)};
  \fill[orange!22] (6.5,0.0) rectangle (12.35,0.5);
  \draw (6.5,0.0) rectangle (12.35,0.5);
  \node at (9.4,0.25) {surviving top of lo, doubled (guard at $2^{1}$)};
  \fill[red!18] (12.35,0.0) rectangle (13.2,0.5);
  \draw (12.35,0.0) rectangle (13.2,0.5);
  \node at (12.78,0.25) {st};
  \fill[orange!10] (6.2,-1.05) rectangle (13.2,-0.6);
  \draw[dashed] (6.2,-1.05) rectangle (13.2,-0.6);
  \node[font=\tiny] at (9.7,-0.82) {bits of lo shifted below the guard (dropped)};
  \draw[->] (13.0,-0.6) .. controls (13.55,-0.35) .. (12.9,-0.02);
  \node[anchor=west, font=\tiny, text width=2.2cm] at (13.35,0.25) {OR-fold $\Rightarrow$\\ sticky bit $2^{0}$};
\end{tikzpicture}
\caption{The two \fma{} datapaths, schematic (not to scale). The exact-wide
reference keeps a field wide enough ($\Delta\le421<470$) that alignment loses
nothing. The sticky-fold keeps the same high significand at weight $2^{49}$
and only the top of the shifted low operand, doubled to expose a guard bit;
everything shifted below the guard is OR-folded into a single sticky bit ---
exactly the information RNE needs (\S\ref{sec:background}). Section~\ref{sec:equiv}
proves the two produce identical results on all $2^{96}$ inputs.}
\label{fig:datapaths}
\end{figure*}
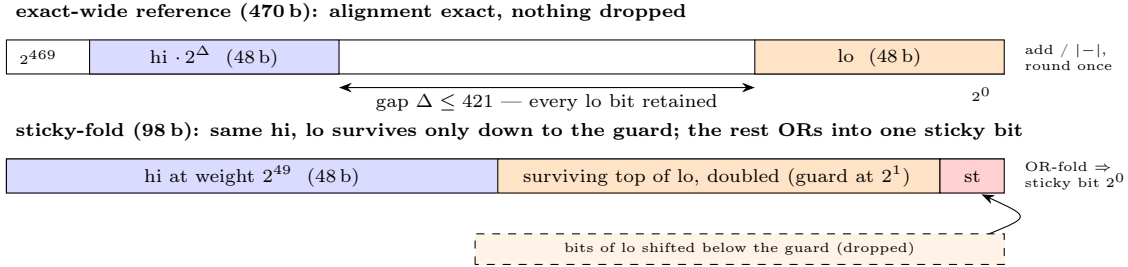

We reimplement the \fma{} as a bounded guard/round/sticky datapath. Both datapaths
share the front end --- unpack to sign / biased exponent / $24$-bit significand
(subnormals floored at exponent $-149$), form the $48$-bit product, select the
higher operand by unbiased exponent. Let $\Delta$ be the exponent gap; for finite
operands $\Delta \le 421$.

\paragraph{Reference (\code{arch_fma_f32_ref}).} Zero-extend both significands to
$\mathrm{FMA\_W}=470$ bits, shift the higher one up by $\Delta$, add or take the
absolute difference, round once. Since $\Delta \le 421 < 470$ the field is
exact; that the result is $\rne(\text{exact }a\cdot b + c)$ is itself a
theorem, not a construction argument: the aligned magnitude and selected sign
are proved equal to the decoded $|a\cdot b + c|$ and its sign
(\code{fma_mag_exact}, \code{fma_sign_exact}; Section~\ref{sec:trust}). This
is the specification (and the $160$-MHz circuit above that our flow could not
retime).

\paragraph{Sticky-fold (\code{arch_fma_f32}).} Use a $98$-bit field
(Figure~\ref{fig:datapaths}) with
$\mathrm{FMA\_G}=48$ guard bits: the high significand sits at weight $2^{49}$ and
the shifted low significand is doubled to make room for a guard bit, with one
sticky bit absorbing everything the shift drops. The guard width
$\mathrm{FMA\_G}=48$ is chosen so that any cancellation in $a\cdot b + c$ that could
affect rounding keeps the product/addend least-bit gap at most $47$ --- it stays
\emph{inside} the guard region, so the fold below it loses no rounding
information. This is the structural fact the equivalence proof formalizes.

Being a fixed-width GRS datapath rather than a $470$-bit adder, the sticky-fold
pipelines. Table~\ref{tab:fmasweep} reports an $f_{\max}$-versus-pipeline-depth
sweep on the same Nangate45 cells and flow: $180$\,MHz combinational, rising to a
$260$--$268$\,MHz plateau at five to seven stages, with the best measured point at
seven stages ($268$\,MHz). Beyond the plateau more registers only add fixed
per-stage overhead (clock-to-Q $+$ setup $\approx 0.4$\,ns each) without splitting
the remaining fine-grained cones --- eight and ten stages regress. The
depth--$f_{\max}$ relation is also non-monotone within the sweep because register
\emph{placement} strategy matters as much as count.

\begin{table*}[t]
\centering
\begin{tabular}{lrrrrrrrr}
\toprule
pipeline stages & 0 (comb.) & 3 & 4 & 5 & 6 & \textbf{7} & 8 & 10\\
\midrule
$f_{\max}$ (MHz) & 180 & 223 & 223 & 260 & 260 & \textbf{268} & 254 & 242\\
cells & 10805 & 12466 & 12424 & 13224 & 13196 & \textbf{13935} & 14127 & 16236\\
\bottomrule
\end{tabular}
\caption{Bounded sticky-fold FP32 \fma{}: $f_{\max}$ and cell count vs.\ pipeline
depth, Nangate45 standard cells (typical corner), Yosys $+$ OpenSTA, buffered flow
(Remark~\ref{rem:buffering}; best register-placement strategy per depth). The
bounded datapath pipelines cleanly to a $260$--$268$\,MHz plateau --- unlike the
exact-wide reference, which retiming does not improve at all.}
\label{tab:fmasweep}
\end{table*}

Because both tables use the same Nangate45 flow, the numbers are directly
comparable, and they separate the two effects cleanly. \emph{Combinationally} the
sticky-fold is a modest win: $180$\,MHz vs.\ the exact-wide's $160$\,MHz, at
$1.8\times$ less area ($10.8$k vs.\ $19.2$k cells). The decisive difference is
\emph{pipelinability}: the sticky-fold reaches $268$\,MHz at seven stages ---
$1.68\times$ the exact-wide's ceiling, since the exact-wide gained nothing from
retiming in our flow. The \fma{} was the only operator in
Table~\ref{tab:peropfmax} whose ceiling demanded a redesign; at seven stages it
sits within $3\%$ of FP32 \code{mul} ($268$ vs.\ $276$\,MHz) instead of
$40\%$ below the field.

Replacing a verified datapath with a faster one creates a new obligation: does the
fold change the result? The next section answers it.

\section{Bit-exact equivalence of the two \fma{} datapaths}\label{sec:equiv}

Our central new theorem is a \emph{sorry}-free Lean~4 proof
\begin{multline}\label{eq:main}
  \forall\, a,b,c \in \mathrm{BitVec}\,32,\\
  \code{arch_fma_f32}(a,b,c) = \code{arch_fma_f32_ref}(a,b,c),
\end{multline}
over the entire $2^{96}$ input space. Composed with the reference's proved
correct rounding (Section~\ref{sec:trust}), it yields the machine-checked
end-to-end theorem \code{arch_fma_f32_correct}: the synthesizable bounded
\fma{} is correctly rounded against the exact value of its decoded operands
--- proved without ever re-engaging the multiplier.

\paragraph{Why it is tractable.} Both datapaths compute the \emph{same} product
significand and apply the \emph{same} special-value wrapper; in the
equivalence miter these cancel under common-subexpression elimination, so neither
a bit-blaster nor the Lean proof ever solves a multiplier equivalence. The
residual content is (a) bit-precise field construction and operand selection,
discharged by \code{bv_decide} over concrete widths --- an \emph{identity} over the
shared multiply, hence cheap --- and (b) rounding, discharged structurally on
integers; the variable alignment shift is never bit-blasted.

\paragraph{Rounding collapse.} Reduce each \fma{} to
$\code{roundNE_f32}(s,m,e)$ and compare magnitudes:

\begin{lemma}[GRS collapse]\label{lem:collapse}
Let $m_1,m_2$ agree above bit $g$ (\,$m_1/2^g = m_2/2^g$\,) and share their low-$g$
sticky status (\,$m_1 \bmod 2^g = 0 \iff m_2 \bmod 2^g = 0$\,). If $g$ lies at or
below the rounding position --- $g+24 \le \log_2 m_1$ for a normal result, or
$g < -149-e$ for a subnormal one --- then
$\code{roundNE_f32}(s,m_1,e) = \code{roundNE_f32}(s,m_2,e)$.
\end{lemma}

To apply Lemma~\ref{lem:collapse} across datapaths, the proof characterizes the
sticky-fold and reference magnitudes as closed-form integer expressions (for both
same- and opposite-sign alignment), relates them by an exact power of two when
$\Delta\le48$, and relates the two alignment exponents through a balanced-modulus
bridge at the signed-integer level; an exponent-alignment lemma scales the
sticky-fold magnitude into the reference's frame.

\paragraph{The case lattice.} The top theorem is a decision tree: NaN / infinity /
zero-factor special cases (each a two-step rewrite chaining a sticky-fold
special-value lemma with its reference mirror, both discharged by \code{bv_decide}
on the $98$- and $470$-bit datapaths); zero addend; exact cancellation
($\text{magnitude}=0\Rightarrow\pm0$); and the finite non-cancelling region, split
by sign, by $\Delta\le48$ vs.\ $\Delta>48$, and by the higher significand's
magnitude (with the normal / subnormal result branch dispatched internally via
Lemma~\ref{lem:collapse}).

\paragraph{The hard case.} One configuration resists the generic collapse: an
opposite-sign \fma{} whose higher significand is exactly a power of two, with a
non-zero lower operand and a normal result. The subtraction borrow drops the
leading bit, so the round shift coincides with the natural collapse position and
violates the strict precondition of Lemma~\ref{lem:collapse}. We resolve it in two
pieces: at $\Delta=49$ the fold is the identity, so the two magnitudes are equal
and the rounding is trivial; for $\Delta\ge50$ a tighter bound on the folded low
part lets the collapse be taken one bit lower, satisfying the precondition with
equality. This case is a good illustration of why the development must be
bit-precise: the rounding decision turns on a single guard bit at a position that
moves with the inputs.

\section{Trust base, audit, and assurance}\label{sec:trust}

\paragraph{Axioms.} \texttt{\#print axioms} on the equivalence and end-to-end
theorems reports the standard three (\code{propext}, \code{Classical.choice},
\code{Quot.sound}) plus one
\texttt{\_native.}\allowbreak\texttt{bv\_decide.}\allowbreak\texttt{ax\_*}
axiom per \code{bv_decide} invocation. To be precise about what those axioms
mean: the external SAT solver (CaDiCaL) is \emph{untrusted} --- it must emit an
LRAT refutation certificate~\cite{lrat} (a SAT-unsatisfiability proof format
whose steps carry explicit propagation hints, checkable in linear time without
search --- which is what makes a small \emph{verified} checker practical),
validated by exactly such a formally verified checker --- but the checker is
executed by Lean's \emph{compiled native evaluator}, and
its verdict enters the proof as an axiom. The trusted base for the
\code{bv_decide} steps is therefore the Lean kernel \emph{plus} the Lean
compiler and native-evaluation bridge (the same trust class as
\code{native_decide}), not the kernel alone~\cite{bvdecide}. By contrast, the
value-level rounder anchor is kernel-only: \texttt{\#print axioms} on
\code{roundNE_nearest} reports exactly the standard three (and on the R1 kernel
characterization \code{rneQuot_nearest}, only \code{propext} and
\code{Quot.sound}) --- no native evaluation anywhere in that chain; the native
trust enters solely through the bit-level datapath lemmas. There is no
\code{sorry}, no hand-declared axiom, and no soundness-weakening option. The
development is about $9\,000$ lines across $23$ modules (Lean v4.30.0).

\paragraph{Artifact.} Every machine-checked claim in this paper --- the Lean
theorems, the Tier-1 SMT proofs, the renderer miters, and the archived
synthesis flow --- is checked against one release of the ARCH repository:
\code{v0.71.0} (commit \texttt{1a7d9fd}),
\url{https://github.com/arch-hdl-lang/arch-com/releases/tag/v0.71.0}.

\paragraph{Independent audit.} An adversarial audit, run separately, rebuilt the
project, re-derived the axiom set, and --- the strongest check against a hollow
proof --- \emph{regenerated} the Lean model from the operator source and confirmed
it byte-identical to the committed model. So \code{arch_fma_f32} and
\code{arch_fma_f32_ref} are the genuine emitted artifacts, structurally
independent ($98$-bit fold vs.\ $470$-bit exact), hence not trivially equal; the
audit further confirmed the top-level case split exhaustive and non-vacuous, every
\code{bv_decide} operating on the real datapaths, and the rounder a genuine RNE
encoder.

\paragraph{Where the IEEE anchor bottoms out.} No verification proves a
specification correct absolutely; it bottoms out in a trusted definition. For
\fma{} that definition is $\rne(\text{exact }a\cdot b+c)$ --- and the chain
from it to the emitted datapath is now proved \emph{end to end}, with no
by-construction step. A value-level Lean theorem
(\code{arch_fma_f32_correct}, module \code{ArchFpEquiv.FmaValue}) states that for finite nonzero operands whose
exact result is nonzero and in range, the \fma{} returns the finite FP32
pattern nearest in magnitude to the exact real value $a\cdot b+c$ among
\emph{all} finite patterns, with ties to an even mantissa
(\code{roundNE_tie_even}), the exact sign, and overflow to signed infinity
occurring exactly when the correctly rounded magnitude reaches $2^{128}$, the
IEEE criterion (\code{arch_fma_f32_overflow},
\code{biasedFinal_overflow_iff}). Both halves of the specification are
theorems rather than construction arguments: ``exact'' is the magnitude
identity \code{fma_mag_exact} --- the reference's aligned $470$-bit magnitude
at its alignment exponent equals $|a\cdot b+c|$ of the decoded operands, with
the datapath's sign selection equal to the exact value's sign
(\code{fma_sign_exact}) --- and ``RNE'' is the rounder theorem
\code{roundNE_nearest}. All quantities in play are dyadic rationals, so every
comparison is carried out \emph{exactly} over $\mathbb{N}$ (the value
$a\cdot b+c$ lives on the fixed grid $2^{-298}$); no reals and no rational
arithmetic are needed, and the proofs stay in core Lean. The trusted base
thus shrinks to three defining lines a reviewer checks by reading: the signed
magnitude an FP32 bit pattern denotes (in units of $2^{-149}$), the exact
value $\code{fmaExact} = s_A s_B + s_C\cdot 2^{149}$ (in units of
$2^{-298}$), and the nearest-among-representables predicate. The datapath
additionally remains cross-checked by the SMT \texttt{FloatingPoint} theory
and the SoftFloat differential.

\section{Related work}\label{sec:related}

Mechanized floating-point verification has a long lineage: Harrison verified IEEE
division and transcendental functions for Intel in HOL Light~\cite{harrison};
Russinoff verified arithmetic units at AMD in ACL2~\cite{russinoff}; the
Flocq~\cite{flocq} Coq library formalizes IEEE-754 and underpins verified
compilers and the FloVer~\cite{flover} round-off analyzer. These efforts prove
hardware or kernels correctly rounded against a real/rational specification --- the
methodology our Tier-2 multiplier proofs follow, through to the end-to-end
value-level \fma{} theorem. On the automated side, the SMT-LIB
\texttt{FloatingPoint} theory~\cite{smtfp} and solvers such as z3, cvc5, and
Bitwuzla decide bit-vector floating-point formulas, and Lean's
\code{bv_decide}~\cite{bvdecide} brings LRAT-certified bit-blasting into
Lean, validating solver certificates with a verified checker run under compiled
native evaluation; all hit the multiplier wall on \fma{}, which our two-tier
split (exhaustive SMT below the wall, structured Lean above) and our
multiplier-cancelling equivalence proof are organized to route around. Berkeley
SoftFloat~\cite{softfloat} is the standard differential oracle. We are not aware of
prior work that delivers FP32/BF16 \emph{as a verified, synthesizable HDL feature}
from a single source, nor of the specific device of pairing a provably-wide exact
\fma{} reference with a bounded GRS datapath and joining them by an equivalence in
which the multiplier cancels.

\section{Limitations}\label{sec:limitations}

The proofs establish equivalence and correct rounding of the Lean-rendered models;
transfer to the emitted SystemVerilog rests on the SV and Lean renderers denoting
the same IR semantics. The shared linearizer reduces this to a per-operator
correspondence rather than a whole-function obligation, and for the
SystemVerilog renderer that correspondence is now machine-checked: Yosys ---
an independent implementation of SystemVerilog semantics --- reads the
emitted SV and exports it to SMT, and a solver proves the export equivalent
to the SMT renderer's model, \texttt{unsat} for all $24$ exposed operators
--- including the saturating integer conversions --- (the two
\fma{} miters via a sound $510$-way case-split on the alignment gap, under
which both barrel shifters collapse; the split covers all inputs by
construction). Three mechanical transformations sit inside this path and are
part of its trust base, disclosed in the artifact: a syntactic
declaration-hoisting preprocessor on the SV (Yosys cannot parse
declaration-with-initializer inside functions), Yosys's \code{opt_clean}
dead-wire removal after flattening, and a specialization of the SMT export
that eliminates Yosys's state sort by instantiating it at the single state
constant. Modulo those, a divergence would have to be a bug shared with
Yosys's independent frontend. The Lean renderer remains covered by the
byte-identical regeneration audit. Relatedly, the IR and both formal models are two-state:
SystemVerilog \code{X}/\code{Z} behavior is out of scope, the verified
statements quantify over binary inputs, and driving every input is the
environment's obligation. The IEEE-754 anchor is proved end to end against a value-level
dyadic specification of the exact $a\cdot b+c$ (Section~\ref{sec:trust});
the residual trust there is the three-line spec itself --- the signed
magnitude a bit pattern denotes, the exact-value expression, and the
nearest-among-representables predicate --- checked by reading. The verified
contract is value semantics under static rounding (Section~\ref{sec:verif}):
dynamic rounding modes, IEEE status flags, and signaling comparisons are not
implemented, and NaN payloads follow an explicit canonical profile. The synthesis
numbers are preliminary characterizations from a single flow on a single open
cell library and corner (Nangate45, typical); a multi-corner, multi-library
study is future work. Finally, the BF16 \fma{}
is, by design, fused f32-accumulate and not correctly rounded, as
Section~\ref{sec:verif} states.

\section{Conclusion}\label{sec:conclusion}

We added first-class FP32 and BF16 to a hardware description language and
verified every operator against its stated contract from a single source ---
end to end, with the SystemVerilog renderer correspondence machine-checked
and the Lean renderer audited (Section~\ref{sec:limitations}); for the BF16 \fma{}, that contract is the
FP32-accumulating fusion of Section~\ref{sec:bf16fma}. The methodology splits at the
solver-tractability frontier --- exhaustive SMT below the multiplier wall,
structured Lean above --- and the headline result turns a timing problem into a
verification one and dispatches it cleanly: a correctly rounded \fma{} can be made
fast and pipelineable by replacing its exact-wide datapath with a bounded sticky
fold, and the replacement can be proved bit-identical to the exact reference over
all $2^{96}$ inputs \emph{without ever solving a multiplier equivalence}, because
the shared multiplier cancels. We see the two-reference pattern --- an inspectable
wide specification plus a synthesizable bounded implementation, joined by a
multiplier-cancelling equivalence proof --- as broadly applicable to
multiplier-bearing arithmetic that defeats direct bit-blasting.

\paragraph{Availability.} The compiler, the operator IR and its three renderers,
and the Lean development are intended for open release; the proofs are reproducible
with Lean~4 (v4.30.0) and a standard \code{bv_decide} toolchain.


\begin{thebibliography}{99}
\bibitem{bvdecide}
Lean \texttt{bv\_decide}: an LRAT-certified bit-vector decision procedure for
Lean~4 (Lean FRO); certificates are validated by a formally verified checker
executed via Lean's native evaluator.
\end{thebibliography}
\end{document}